\documentclass{article}

\usepackage[preprint]{neurips_2021}

\usepackage[utf8]{inputenc} 
\usepackage[T1]{fontenc}    
\usepackage{hyperref}       
\usepackage{url}            
\usepackage{booktabs}       
\usepackage{amsfonts}       
\usepackage{nicefrac}       
\usepackage{microtype}      
\usepackage{xcolor}         

\usepackage{smile}
\usepackage{svg}
\usepackage{amssymb}

\usepackage{tabularx, booktabs}
\newcolumntype{Y}{>{\centering\arraybackslash}X}
\newcolumntype{L}{>{\arraybackslash}X}

\newcommand{\ie}{\textit{i.e.}}

\newcommand{\smallsection}[1]{{\vspace{0.2cm}\noindent\textbf{#1.}}}

\newcommand{\LN}{{f_{\mbox{\scriptsize LN}}}}
\newcommand{\ATT}{{f_{\mbox{\scriptsize ATT}}}}
\newcommand{\DOT}{{f_{\mbox{\scriptsize Scaled Dot-Product Attention}}}}
\newcommand{\FFN}{{f_{\mbox{\scriptsize FFN}}}}
\newcommand{\softmax}{\mbox{softmax}}
\newcommand{\head}{\mbox{head}}

\newcommand{\wf}{W^{(1)}}
\newcommand{\wff}{W^{(2)}}
\newcommand{\wk}{W^{(K)}}
\newcommand{\wv}{W^{(V)}}
\newcommand{\wq}{W^{(Q)}}
\newcommand{\wo}{W^{(O)}}

\title{Multi-head or Single-head? 
An Empirical Comparison for Transformer Training}

\author{Liyuan Liu \\
  University of Illinois at Urbana-Champaign\\
  \texttt{ll2@illinois.edu} \\
  \And
  Jialu Liu\\
  Google Research\\
  \texttt{jialu@google.com}\\
  \And 
  Jiawei Han\\
  University of Illinois at Urbana-Champaign\\
  \texttt{hanj@illinois.edu}\\
}

\begin{document}

\maketitle

\begin{abstract}
Multi-head attention plays a crucial role in the recent success of Transformer models, which leads to consistent performance improvements over conventional attention in various applications. 
The popular belief is that this effectiveness stems from the ability of jointly attending multiple positions. 
In this paper, we first demonstrate that jointly attending multiple positions is not a unique feature of \emph{multi-head} attention, as \emph{multi-layer single-head} attention also attends multiple positions and is more effective. 
Then, we suggest the main advantage of the multi-head attention is the training stability, since it has less number of layers than the single-head attention, when attending the same number of positions. 
For example, \emph{24-layer 16-head} Transformer (BERT-large) and \emph{384-layer single-head} Transformer has the same total attention head number and roughly the same model size, while the multi-head one is significantly shallower. 
Meanwhile, we show that, with recent advances in deep learning, we can successfully stabilize the training of the 384-layer Transformer. 
As the training difficulty is no longer a bottleneck, substantially deeper single-head Transformer achieves consistent performance improvements without tuning hyper-parameters.

\end{abstract}
\section{Introduction}

Transformers~\citep{Vaswani2017AttentionIA} have led to a series of breakthroughs in various deep learning tasks~\citep{Devlin2019BERTPO,Velickovic2017GraphAN}. 
One distinguishing characteristic of Transformer is that it does not contain any recurrent connections and can parallelize all computations in the same layer, thus leads to better effectiveness, efficiency, and scalability. 
Without using recurrent connections, Transformer purely relies on the attention mechanism to capture the dependency among input tokens. 
Specifically, a novel multi-head attention module was proposed and used in Transformer to better capture the dependency among input tokens. 

This multi-head attention module has been observed to be one major reason behind the success of the Transformer. 
For example, on machine translation benchmarks, Recurrent Neural Networks (RNNs) can outperform Transformers when both are using the multi-head encoder-decoder attention, and would underperform without using the multi-head attention~\citep{Chen2018TheBO}. 
Besides Transformer, multi-head attention has also been incorporated into other models like RNNs~\citep{Chen2018TheBO}, Graph Attention Network~\citep{Velickovic2018GraphAN}, and Convolutional Neural Network~\citep{Xiao2020CNNMHSAAC,Fang2019SelfMA}.

At a high level, it is widely believed that multi-head attention stands out by jointly attending multiple positions, while conventional attention module can only attend one position in one layer. 
Specifically, multi-head attention projects the inputs into multiple different subspaces and conduct multiple attention computations in parallel.  

\smallsection{Our Contributions}
Our point of start is demonstrating that attending multiple positions is not a unique feature of multi-head attention. 
In fact, stacking multiple conventional attentions can also attend multiple positions, and even could be more effective than the multi-head attention.

\begin{figure}[t]
\centering
\includegraphics[width=\linewidth]{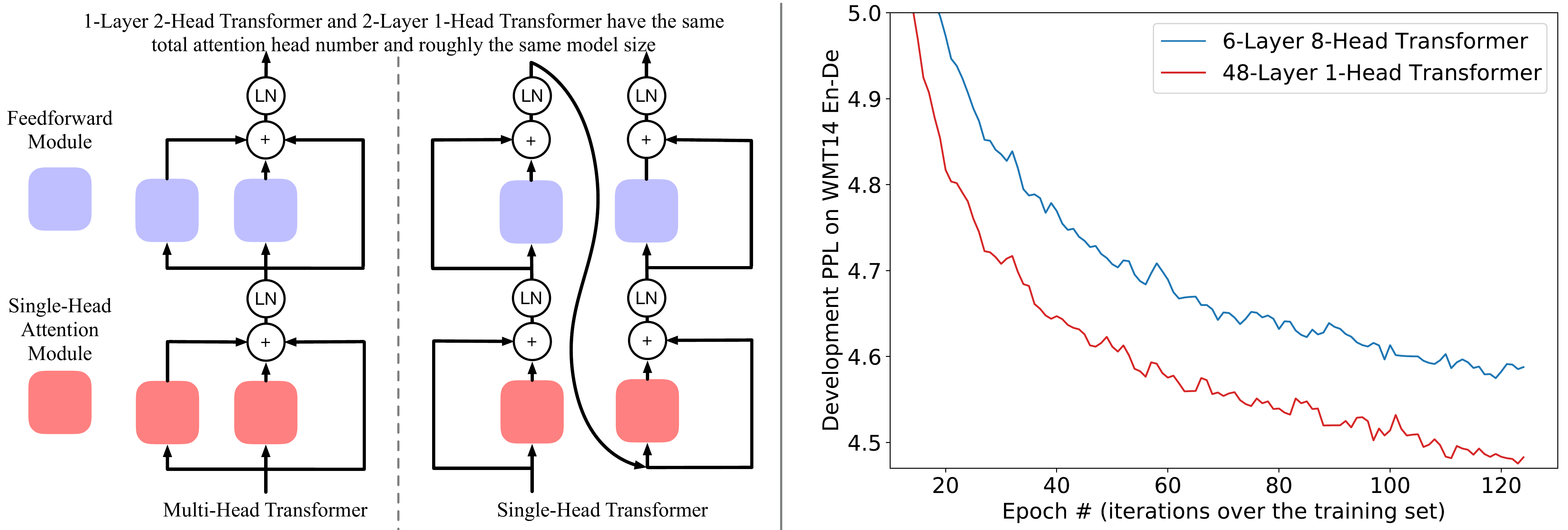}
\caption{Left: with the same model components, both multi-head and multi-layer Transformer can attend multiple positions. Right: comparing to multi-head Transformer, multi-layer Transformer has the potential to achieve a better performance, while its training is more challenging (without Admin, the 48-layer transformer training diverged).}
\label{fig:multi-head-single-head}
\end{figure}

Specifically, as in Figure~\ref{fig:multi-head-single-head}, a multi-head attention module can be viewed as an ensemble model, which combines multiple single-head attention modules by calculating their average.
Thus, by integrating these modules differently, we can reconstruct a Transformer to be single-head\footnote{We use single-head/multi-head Transformer to refer Transformer with single-head/multi-head Attention.} and substantially deeper. 
These two networks can attend the same number of places (i.e., have the same total number of attention heads ), have roughly the same number of parameters and inference computation complexity, while the multi-head one is shallower and the single-head one is deeper. 

In our experiments, we observe that, comparing to the shallower multi-head Transformer, the deeper single-head Transformer is more effective but harder to train, which matches the common wisdom that model depth can increase model capacity at the cost of training difficulty. 
For example, the 6-layer 6-head Transformer encoder-decoder model converges well, while the 36-layer single-head Transformer encoder-decoder model diverges. 
Fortunately, the recent advances in deep learning successfully stabilize the Transformer training, and allows us to train substantially deeper models. 
Specifically, with the adaptive model initialization (Admin), we are able to train the 36-layer single-head Transformer, without changing any hyper-parameters~\citep{Liu2020UnderstandingTD}. 
Meanwhile, after stabilizing the training of the 36-layer model, it converges faster and better than the 6-layer one (as in Figure~\ref{fig:multi-head-single-head}).  
As elaborated in Section~\ref{sec:exp}, we also reconstruct the 12-layer 12-head BERT-base model into 144-layer single-head model, 24-layer 16-head BERT-large into 384-layer single-head model, and find both performs better consistently across various tasks and domains. 

\section{Related Work}

There exist two aspects of related work regarding the
topic here, which are Attention and Transformer. 

\subsection{Attention and Multi-Head Structure}

Attention modules are first proposed to capture the long-term dependency in sequence-to-sequence models~\citep{Graves2014NeuralTM,Bahdanau2015NeuralMT}. 
To calculate the output for a token in the target sequence, the attention module would calculate a weighted average of source token representations, while the weight is calculated by applying softmax on attention scores. 
Different variants of attention modules calculate attention scores differently. 
For example, to calculate the attention score, \citet{Graves2014NeuralTM} uses the cosine similarity, \citet{Bahdanau2015NeuralMT} uses the perception network, and \citet{Luong2015EffectiveAT} uses dot product. 
While these modules can only attend one position in one layer, multi-head attention is developed to improve the conventional attention module by allowing the module jointly attending multiple positions~\citep{Vaswani2017AttentionIA}, which is identified as one major reason behind the success of Transformer~\citep{Chen2018TheBO}. 
Also, it has inspired several follow up work to analysis the multi-head structure~\citep{Michel2019AreSH,Peng2020AMO}. 
Specifically, \citet{Michel2019AreSH} observes single-head Transformer performing better than multi-head Transformer for model pruning. 
Still, no study has been conducted on deep single-head Transformer training, due to its training difficulty. 

\subsection{Transformer}

Transformer~\citep{Vaswani2017AttentionIA} has led to a series of breakthroughs in various domains~\citep{Devlin2019BERTPO,Velickovic2017GraphAN,Huang2019MusicTG,Parmar2018ImageT,Ramachandran2019StandAloneSI}.
Meanwhile, Transformer training has been found to be more challenging and attracted lots of attention to analyze why Transformer is harder to train and how to stabilize Transformer training~\citep{Liu2019OnTV, Baevski2018AdaptiveIR,Nguyen2019TransformersWT,wang-etal-2019-learning, Xiong2019OnLN,Liu2020UnderstandingTD}. 
Many efforts have been made to improve Transformer, e.g., relative position encoding~\citep{Shaw2018SelfAttentionWR} or replacing dot-product attention with locality-sensitive hashing~\citep{Kitaev2020ReformerTE}. 
Here, we choose to focus our study on the original Transformer model as proposed in \citet{Vaswani2017AttentionIA}, and uses the initialization technique Admin to stabilize model training \citep{Liu2020UnderstandingTD}, since this method does not include any additional hyper-parameters and its final model is equivalent to the original Transformer. 
 
\section{Transformer Architecture}
\label{sec:architecture}
The Transformer architecture contains two types of sub-layers, i.e., Attention sub-layers and Feedforward sub-layers. 
Each sub-layer is constructed with the  shortcut connection and the Layer Norm. 
Specifically, it calculates the output as $\xb_{i+1} = \LN(\xb_i + f(\xb_i))$, where $\xb_i$ is the input of layer $i$  and the output of layer $i-1$ (top layers have larger indexes), $\LN$ is the Layer Norm , and $f(\cdot)$ is multi-head attention $\ATT(\cdot)$ or feedforward  $\FFN(\cdot)$ for Attention sub-layers and Feedforward sub-layers respectively. 

\smallsection{Layer Norm}
Layer norm~\citep{Ba2016LayerN} plays a vital role in the Transformer architecture. It is defined as $\LN(\xb) = \bgamma \frac{\xb - \mu }{\sigma} + \bnu$, where $\mu$ and $\sigma$ are the mean and standard deviation of 
$\xb$, $\bgamma$ and $\bnu$ are learnable parameters.

\smallsection{Feedforward}
Transformers use two-layer perceptrons
as feedforward networks,
\ie, $\FFN(\xb) = \phi(\xb \wf)\wff$, where $W^{(\cdot)}$ 
are parameters, and $\phi(\cdot)$ is the non-linear function. Specifically, the original Transformer ReLU as the activation function, while later study uses other types of activation function, e.g., BERT uses GELU as the activation function~\citep{Hendrycks2016GaussianEL}.

\smallsection{Attention}
Transformers use the multi-head attention to capture the dependency among input tokens, which is based on the scaled dot-product attention. 
Scaled dot-product attention tries to query information from the source sequence that is relevant to the target sequence. 
Specifically, assuming the length of the source sequence and the target sequence to be $n$ and hidden dimension to be $m$, target sequence would be encoded as $Q\in R^{n\times m}$, source sequence would be encoded as $K\in R^{n\times m}$ and $V\in R^{n\times m}$. 
The scaled dot-product attention would calculate the output as $\DOT(Q, K, V) = \softmax(\frac{QK^T}{\sqrt{m}})V$, where $\softmax(\cdot)$ is the row-wise softmax. 

One scaled dot-product attention is believed to attend only one position in each row (for each target token), since the output of softmax typically would have one dimension significantly larger than other dimensions in each row.
Multi-head attention was proposed to jointly attend multiple positions, which employs multiple scaled dot-product attention in parallel. 
Specifically, it calculates the output as $\ATT(Q, K, V) = [\head_1; \cdots; \head_h] \wo$, where $\head_i = \DOT(Q\wq_i, K\wk_i, V\wv_i)$, $W^{(\cdot)}$ are learnable parameters, and $h$ is the number of heads.

\smallsection{Transformer}
Transformer has two types of layer configurations when serving as the encoder and the decoder respectively.
Here, we use $\xb_i$ to refer the input of sub-layer $i$. 
Each Transformer encoder layer contains two sub-layers, i.e., one attention sub-layer in a self-attention manner and one feedforward sublayer.
Specifically, the attention sub-layer calculates outputs as $\xb_{2i+1} = \LN(\xb_{2i} + \ATT(\xb_{2i}, \xb_{2i}, \xb_{2i}))$ and the feedforward sub-layer calculates outputs as $\xb_{2i+2} = \LN(\xb_{2i+1} + \FFN(\xb_{2i+1})$. 
Notice that the attention sub-layer sets $Q$, $K$, and $V$ as the same value  $\xb_{2i}$, capturing the dependency among tokens within the same sequence, which is referred as self-attention. 

Each Transformer decoder layer contains three sub-layers, besides the self-attention sublayer and the feedforward sublayer, it also includes a encoder-decoder attention sub-layer between them. 
Specifically, the encoder-decoder attention sub-layer calculates outputs as $\xb_{3i+2} = \LN(\xb_{3i+1}+\ATT(\xb_{3i+1}, \hb, \hb)$, where $K$ and $V$ are set to the encoder output $\hb$. 

\section{From Shallow Multi-Head To Deep Single-Head}

The multi-head attention features the ability to jointly attend multiple positions in the same layer. 
Here, we first show that the multi-head attention sub-layers and the feedforward sub-layers have inherent ensemble structures combining multiple smaller modules (e.g., outputs of 8-head attention is the sum of 8 single-head attention). 
Then, we reconstruct the shallow multi-head Transformer into single-head multi-layer Transformer, which combines these modules in a more effective manner by allowing them to enhance each other. 

\subsection{Inherent Ensemble Structure within Transformer}
\label{sec:parallel}

As in Figure~\ref{fig:multi-head-single-head}, multi-head attention sub-layers and feedforward sub-layers have the inherent ensemble structure, i.e., each of these sub-layers can be viewed as an ensemble of smaller models. 
Now let us proceed to introduce those parallel structure in details. 
Note that notations are introduced in Section~\ref{sec:architecture}. 

\smallsection{Attention}
We split the weight matrix $\wo$ into $h$ parts by rows, i.e., we mark $\wo = [W^{(O)^T}_1; \cdots;W^{(O)^T}_h]^T$. 
Then, the multi-head attention calculates outputs as:
\begin{align}
    \ATT(Q, K, V) &= [\head_1; \cdots; \head_h] \wo \nonumber\\
    &= \sum_{i=1}^h \head_i \wo_i = \sum_{i=1}^h \softmax(\frac{Q\wq_iW^{(K)^T}_jK^T}{\sqrt{m}})V\wv_i\wo_i. \label{eqn:attn}
\end{align}
Note that on the right side of Equation~\ref{eqn:attn}, each term can be viewed as a low-rank version of the general attention~\citep{Luong2015EffectiveAT}. 
Thus, the multi-head attention can be viewed as jointly attending multiple places by ensembling multiple conventional attention modules.  

Specifically, the general attention module~\citep{Luong2015EffectiveAT} calculates outputs as:
\begin{align}
    f_{\mbox{\scriptsize General Attention}} (Q, K, V) = \softmax(QW_1K^T)VW_2 \label{eqn:general_attn}
\end{align}
Comparing Equation~\ref{eqn:general_attn} and the term in Equation~\ref{eqn:attn}, we can find their major difference is that the multi-head attention decomposes the $m \times m$ matrix $W_1$ and $W_2$ into $\frac{\wq_i W^{(K)^T}_i}{\sqrt{m}}$ and $\wv_i W^{(O)}_i$, where $\wq_i, \wk_i, \wv_i, W^{(O)^T}_i \in R^{m \times \frac{m}{h}}$. 
With this low rank decomposition, the parameter number and computation complexity of the multi-head attention module would stay the same no matter what the value of $h$ is (i.e., how many heads one layer has). 

\smallsection{Feedforward}
Similar to the Attention module, we can also rewrite the Feedforward sub-layer as an ensemble of $h$ modules.\footnote{Note $h$ here is decided to be consistent with the Multi-Head Attention sub-layers.} 
Specifically, we split the weight matrix $\wf$ into $h$ parts by rows and $\wff$ into $h$ parts by columns, i.e., we mark $\wf = [\wf_1; \cdots;\wf_h]$ and $\wff = [W^{(2)^T}_1; \cdots;W^{(2)^T}_h]^T$. 
Then, the feedforward sub-layer calculates outputs can be rewrote as:
\begin{align}
    \FFN(\xb) = \phi(\xb \wf)\wff = \sum_{i=1}^h \phi(\xb \wf_i)\wff_i \label{eqn:ffn}
\end{align}
Thus, the Feedforward sub-layer can be viewed as an ensemble of $h$ sub-modules. 
Note that since the sum of the $h$ sub-modules would be normalized by Layer Norm, their outputs are integrated in an averaging manner. 

\smallsection{Average Ensemble}
Each Transformer sub-layer calculates outputs as $\LN(\xb+f(\xb))$, where $f(\cdot)$ could be $\FFN(\cdot)$ and $\ATT(\cdot)$. 
Thus, the sum calculated in Equation~\ref{eqn:attn} and \ref{eqn:ffn} would be normalized by $\Var[\xb + f(\xb)]$. 
In this way, the joint effect of layer norm and the sum would be similar to combining these modules in an average ensemble manner. 

\subsection{Shallow Multi-Head and Deep Single-Head as Module Integration Strategy}

Intuitively, with the same set of modules, no matter how these modules are integrated, the model can attend the same number of places. 
Still, some module integration strategy could be more effective integrating modules. 

In the original multi-head Transformer, modules in the same layer are combined in an ensemble manner and cannot enhance each other. 
For example, as in Figure~\ref{fig:multi-head-single-head}, when constructed in the multi-head manner, the two attention heads would have the same input and are agnostic to each other. 
In this way, the second attention head cannot leverage or benefit from the information captured by the first attention head. 

Intuitively, it could be beneficial to allow the second attention head standing on the shoulders of the first attention head. 
To this end, we integrate these modules differently, and reconstruct the shallow multi-head Transformer into the deep single-head Transformer (As in Figure~\ref{fig:multi-head-single-head}). 
Note that both models have the same total number of attention heads, roughly same model size, and roughly the same inference computation complexity.

\section{Multi-Head or Single-Head? Empirical Comparisons} 
\label{sec:exp}

\begin{table}
\begin{center}
\caption{GLUE task descriptions and statistics. The second and fourth column denotes the number of training examples and the number of classes. Note that STS-B is an regression task. }
\label{tbl:glue_description}
\begin{tabularx}{\linewidth}{lllp{2.5cm}LL}
\toprule
Corpus & $|\mbox{Train}|$ & $|\mbox{Label}|$ & Task & Metric(s) & Domain \\
\midrule
\multicolumn{6}{c}{Single-Sentence Classification}\\
\midrule
CoLA & 8.5k & 2 & acceptibility & Matthews corr. & misc.\\
SST-2 & 67k & 2 & sentiment & accuracy & movie reviews\\
\midrule
\multicolumn{6}{c}{Sentence Similarity/Paraphrase}\\
\midrule
MRPC & 3.7k & 2 & paraphrase & accuracy/F1 & news \\
STS-B & 5.7k & - & similarity &  Pearson/Spearman corr. & misc.\\
QQP & 364k & 2 & similarity & accuracy/F1 & social QA questions \\
\midrule
\multicolumn{6}{c}{Natural Language Inference (NLI)} \\
\midrule
MNLI & 393k & 3 & NLI & (mis)matched acc. & misc. \\
QNLI & 108k & 2 & QA/NLI & accuracy & Wikipedia \\
RTE & 2.5k & 2 & NLI & accuracy & misc. \\
WNLI & 634 & 2 & coreference/NLI & accuracy & fiction books\\

\bottomrule
\end{tabularx}
\end{center}
\end{table}

Here, we conduct systematic empirical studies to compare the shallow multi-head attention and the deep single-head attention. 
We first show that, although the deep single-head attention is harder to train, the training difficulty is no longer an obstacle, when the recent advances of deep learning is employed to stabilize model training. 
Then, we show that, the deep single-head attention also attends multiple positions and is more effective than the shallow multi-head attention. 
More analyses are further conducted to verify our intuition. 

\subsection{Experiment Setup}

\smallsection{Transformer Model Specifics}
We conduct experiments with three Transformer models, i.e., Transformer-base for the WMT'14 EN-DE translation task, BERT-base and BERT-large for the language model pre-training. 
Specifically, for machine translation, the original Transformer-base model is 8H-6L-6L\footnote{We use ``$\gamma$H-$\alpha$L(-$\beta$L)" to denote that a model has $\gamma$-head $\alpha$-layer encoder and $\gamma$-head $\beta$-layer decoder. } Transformer encoder-decoder with 512-dimension word embedding, 64-dimension per-head attention output, and 2048-dimension feedforward network~\citep{Vaswani2017AttentionIA}.
Here, we compare it with 1H-48L-48L Transformer encoder-decoder with  512-dimension word embedding, 64-dimension per-head attention output, and 256-dimension feedforward network. 
For language model pre-training, BERT-base model is 12H-12L Transformer encoder with 768-dimension word embedding, 64-dimension per-head attention output, and 3072-dimension feedforward network; BERT-large model is 16H-24L Transformer encoder with 1024-dimension word embedding, 64-dimension per-head attention output, and 4096-dimension feedforward network~\citep{Devlin2019BERTPO}. 
Here, we compare them with deep single-head BERT-base model (1H-144L Transformer encoder with 768-dimension word embedding, single-head 64-dimension per-head attention output, and 256-dimension word embedding) and deep single-head BERT-large model (1H-384L Transformer encoder with 768-dimension word embedding, 64-dimension per-head attention output, and 256-dimension word embedding). 
To stabilize 1H-48L-48L Transformer-base and 1H-384L BERT-large, we use the Admin initialization~\citep{Liu2020UnderstandingTD}.

\begin{table}[t]
\begin{center}
\caption{Deep single-head Transformer is harder to train than the shallow multi-head Transformer. }
\label{tbl:stable}
\begin{tabularx}{\linewidth}{lcccccc}
\toprule
 & \multicolumn{2}{c}{Transformer-base} & \multicolumn{2}{c}{BERT-base} & \multicolumn{2}{c}{BERT-large} \\
& 8H-6L-6L & 1H-48L-48L & 12H-12L & 1H-144L & 16H-24L & 1H-384L \\ 
\midrule
Training & \checkmark & $\times$ /\checkmark (w. Admin) & \checkmark & \checkmark & \checkmark & $\times$/\checkmark (w. Admin) \\
\bottomrule
\end{tabularx}
\end{center}
\end{table}

\begin{table}[t]
\begin{center}
\caption{The model performance on the WMT'14 EN-DE dataset. }
\label{tbl:nmt}  
\scalebox{0.9}{
\begin{tabularx}{1.11\linewidth}{cc|cc|cc|YY|YY}
\toprule
\multicolumn{2}{c|}{8H-6L-6L} & \multicolumn{2}{c|}{1H-48L-48L} & \multicolumn{2}{c|}{Evolved{\tiny \citep{so2019evolved}}} & \multicolumn{2}{c|}{{\small 2D-CSANs}{\tiny\citep{yang2019convolutional}}} &
\multicolumn{2}{c}{{\small DARTSformer}{\tiny\citep{Zhao2021MemoryEfficientDT}}}\\ 
\small BLEU & \small Param. & \small BLEU & \small Param. & \small BLEU & \small Param. & \small BLEU & \small Param. & \small BLEU & \small Param. \\
\midrule
27.90 & 63.2M & 28.40 & 63.6M & 28.4 & 64.1M & 28.18 & 88.0M & 28.4 & 65.2M \\
\bottomrule
\end{tabularx}
}
\end{center}
\end{table}

\begin{table}[t!]
    \centering
    \caption{The model performance on dev sets of MNLI and SQuAD 2.0. 
    The FLOPs are calculated for the inference computation of one 512-length input sequence.}
    \label{tbl:mnli_squad}
    \label{tbl:squad}      
    \scalebox{0.98}{
      \begin{tabular}{l | c | c | cc | cc}
        \toprule
         & \multirow{2}{*}{\textbf{FLOPs \#}} & \multirow{2}{*}{\textbf{Param. \#}} & \multicolumn{2}{c|}{\textbf{MNLI Acc.}} &  \multicolumn{2}{c}{\textbf{SQuAD v2.0}}\\
         & & & \textbf{\small match} & \textbf{\small mis-match}  & \textbf{\small exact match} & \textbf{\small F1} \\
        \midrule
        \textbf{12H-12L} {\small BERT}\textsubscript{BASE} & 46.3B & 109.5M & 84.4 & 84.4 & 77.4 & 80.4  \\
        \textbf{1H-144L} {\small BERT}\textsubscript{BASE default} & 46.9B & 110.0M & 85.6 & 85.1 & 79.6 & 82.4  \\
        \textbf{1H-144L} {\small BERT}\textsubscript{BASE Admin} & 46.9B & 110.0M & 85.2 &	85.4 &	79.2 & 82.5 \\
        \midrule
        \textbf{16H-24L} {\small BERT}\textsubscript{LARGE} &  161.8B & 335.1M & 86.3 & 86.4 & 81.0 & 84.3\\
        \textbf{1H-384L} {\small BERT}\textsubscript{LARGE Admin} & 164.1B & 337.4M & 87.7 & 87.5 & 82.6 & 85.7 \\
        \bottomrule        
    \end{tabular}%
    }
\end{table}

\smallsection{Translation}
Here, we conduct experiments on WMT'14 EN-DE and evaluate model performance based on their BLEU score on the test set and perplexity score on the development set. All hyper-parameter settings are adopted from \citet{Liu2020UnderstandingTD}, and are elaborated in the appendix.

\smallsection{BERT}
Here, we follow the training setting from \citet{Devlin2019BERTPO} and evaluate pre-trained language models on the SQuAD 2.0~\citep{rajpurkar2018know} datasets for question answering, and the GLUE benchmark~\citep{wang2018glue}, which includes 9 subtasks (as in Table~\ref{tbl:glue_description}). 
More detailed experiment settings can be found in the appendix for reproduction.

\subsection{Stability Comparison}
\label{subsec:stability}

As in Table~\ref{tbl:stable}, after changing the shallow multi-head Transformer to the deep single-head Transformer, the training fails to converge well for 2 out of 3 models.
Note that, although the 1H-144L BERT-base model converges successfully, the model is sensitive to the choice of initialization. 
Specifically, the BERT-base model and BERT-large model are initialized with truncated normal distribution with 0.02 variance, instead of following the common practice (e.g., using the Kaiming initialization~\citep{He2015DelvingDI} or the Xavier initialization~\citep{Glorot2010UnderstandingTD}).  
We observe that after changing the variance of the initialization, or following the common practice, the training of the 1H-144L BERT-base model would also fail. 

Meanwhile, we show that, with the recent advances in deep learning, the training can be successfully stabilized by Adaptive Model Initialization (Admin), without changing any hyper-parameters~\cite{Liu2020UnderstandingTD}. 
Also, after employing the Admin initializatioin, the 1H-144L BERT-base model can be trained successfully when following the standard Xavier initialization. 
This shows that, although the deep single-head Transformer is harder to train, the training difficulty is no longer an obstacle.

\begin{table}[t]
    \centering
    \caption{The test performance on the GLUE benchmark with metrics described in Table~\ref{tbl:glue_description}.}
    \label{tbl:glue}
    \scalebox{0.83}{
      \begin{tabular}{l|c|ccccccccc}
        \toprule
         & \textbf{\small GLUE}  
         &  \textbf{\small CoLA} & \textbf{\small SST-2} & \textbf{\small MRPC} & \textbf{\small SST-B} & \textbf{\small QQP} & \textbf{\small MNLI-m/mm} & \textbf{\small QNLI}  & \textbf{\small RTE} & \textbf{\small WNLI} \\
        \midrule
        \textbf{\small {12H-12L}} & 78.3 &52.1 & 93.5 & 88.9/84.8 & \textbf{87.1/85.8} & \textbf{71.2/89.2} & 84.6/83.4 & 90.5 & 66.4 & \textbf{65.1} \\
        \textbf{\small {1H-144L}} & 79.4 &\textbf{59.2} & \textbf{94.2} & \textbf{89.3/85.4} & 84.3/83.5 & 70.9/88.9 & \textbf{85.1/84.3} & \textbf{91.0} & \textbf{69.0} & \textbf{65.1} \\
        \midrule
        \textbf{\small {16H-24L}} & 80.5 & 60.5 & 94.9 & 89.3/85.4 & \textbf{87.6/86.5} & \textbf{72.1/89.3} & 86.7/85.9 & 92.7 & 70.1 & \textbf{65.1}  \\
        \textbf{\small {1H-384L}} & 81.3 & \textbf{62.7} & \textbf{95.1} & \textbf{90.5/87.2} & 86.9/86.3 & 71.3/89.1 & \textbf{87.4/86.5} & \textbf{93.3} & \textbf{72.7} & \textbf{65.1}  \\
        \bottomrule        
    \end{tabular}%
    }
\end{table}

\begin{figure}[t]
\centering
\includegraphics[width=\linewidth]{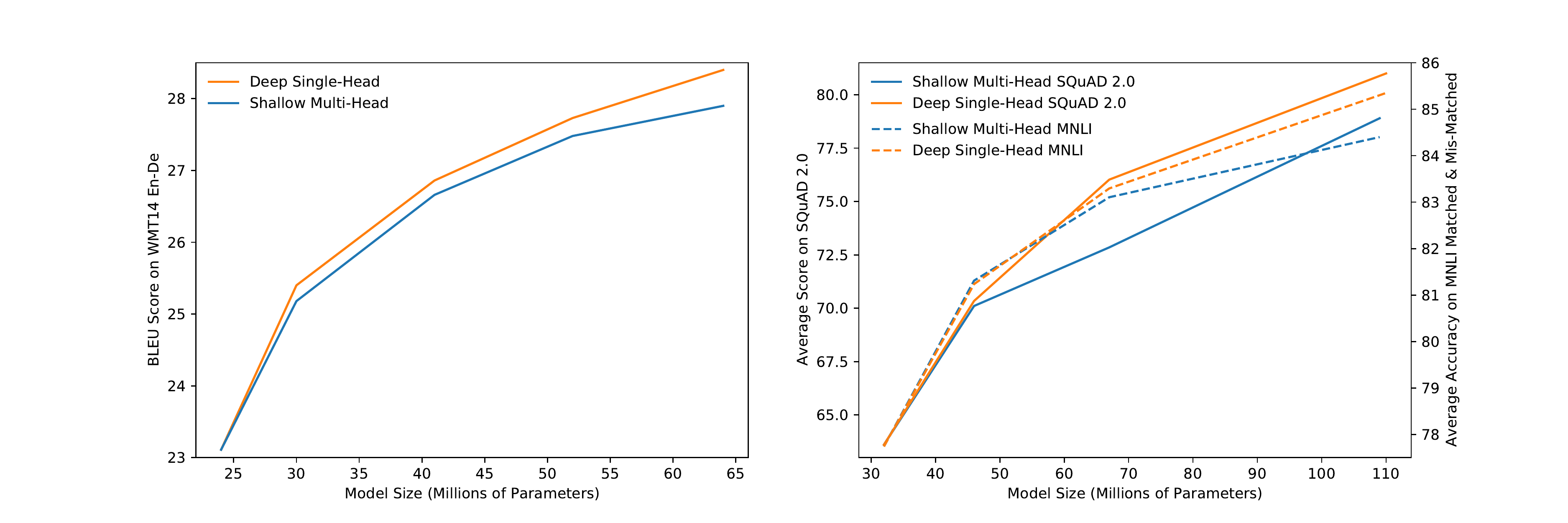}
\caption{Performance Improvements under Different Model Size. Left: the performance of $\alpha$H-6L-6L ($\alpha$=1, 2, 4, 6, 8) and 1H-$\beta$L-$\beta$L ($\beta$=6,12,24,36,48), whose per-head dimension is the same with Transformer-base. Right: the performance of $\alpha$H-12L ($\alpha$=1, 3, 6, 12) and 1H-$\beta$L ($\beta$=12,36,72,144), whose per-head dimension is the same with BERT-base. }
\label{fig:model_size}
\end{figure}

\subsection{Performance Comparison}

For machine translation, we summarize the model performance in Table~\ref{tbl:nmt}. 
With the same model size, the deep single-head Transformer (1H-48L-48L) outperforms the shallow multi-head Transformer. 
Also, the deep single-head Transformer achieves the same performance with the architecture search algorithm the Evolved Transformer~\citep{so2019evolved} and DARTSformer~\citep{Zhao2021MemoryEfficientDT}, with the same parameter number. 
Specifically, Evolved Transformer employs the evolution algorithm to do neural architecture search on Transformer and DARTSformer employs the differentiable neural architecture search algorithm, and both are based on multi-head attention. 
Our deep single-head Transformer achieves similar performance without hyper-parameter tuning, which further verifies its effectiveness. 

We summarized the model performance on the MNLI and SQuAD 2.0 in Table~\ref{tbl:mnli_squad}. 
Similar to machine translation, the deep single-head Transformer achieves consistent performance improvements over the original shallow multi-head Transformer. 
Table~\ref{tbl:glue} shows the test performance on the GLUE benchmark.
The deep single-head Transformer outperforms the shallow multi-head Transformer on 7 out of 9 tasks, and improves the average score (GLUE) by roughly 1 point. 
In the mean time, it is worth mentioning that, on 2 out of 3 sentence similarity/paraphrase tasks, the shallow multi-head Transformer achieves better performance. 
This indicates the deep single-head Transformer can be further improved, and we will further explore this in the future work. 

These observations verified our intuition that the deep single-head Transformer could be more effective than the shallow multi-head Transformer. 
Together with previous analysis in Section~\ref{subsec:stability}, we suggest that the main benefit of the multi-head structure is the improved stability.


\subsection{Impact of Model Initialization}
Previously, we showed that the initialization plays an important role in stabilizing model training (as in Table~\ref{tbl:stable}). 
Here, we aim to understand the impact of model initialization on the model performance. 
Specifically, as the 1H-144L BERT-base model converges well with both the vanilla initialization and the Admin initialization, we not only conduct training with the Admin initialization, but also the vanilla initialization. 
As summarized in Table~\ref{tbl:glue}, the default initialization and the Admin initialization achieves similar performance. 
This observation supports our intuition that the major benefits of the Admin initialization is on training stability, and the performance improvements mostly come from the change from shallow multi-head Transformer to deep single-head Transformer. 

\begin{figure}[t]
\centering
\includegraphics[width=\linewidth]{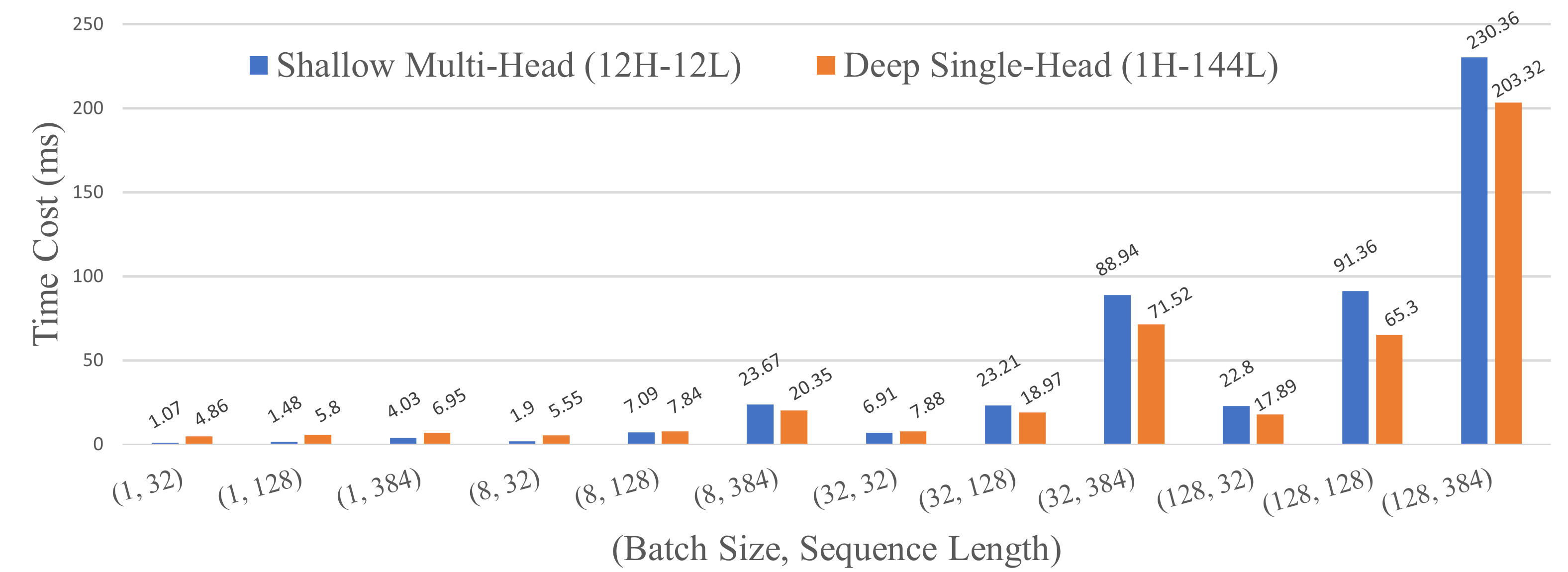}
\caption{The Inference Speed of BERT-base with Different Batch Size and Sequence Length.}
\label{fig:inference-speed}
\end{figure}

\begin{figure}[t]
\centering
\includegraphics[width=\linewidth]{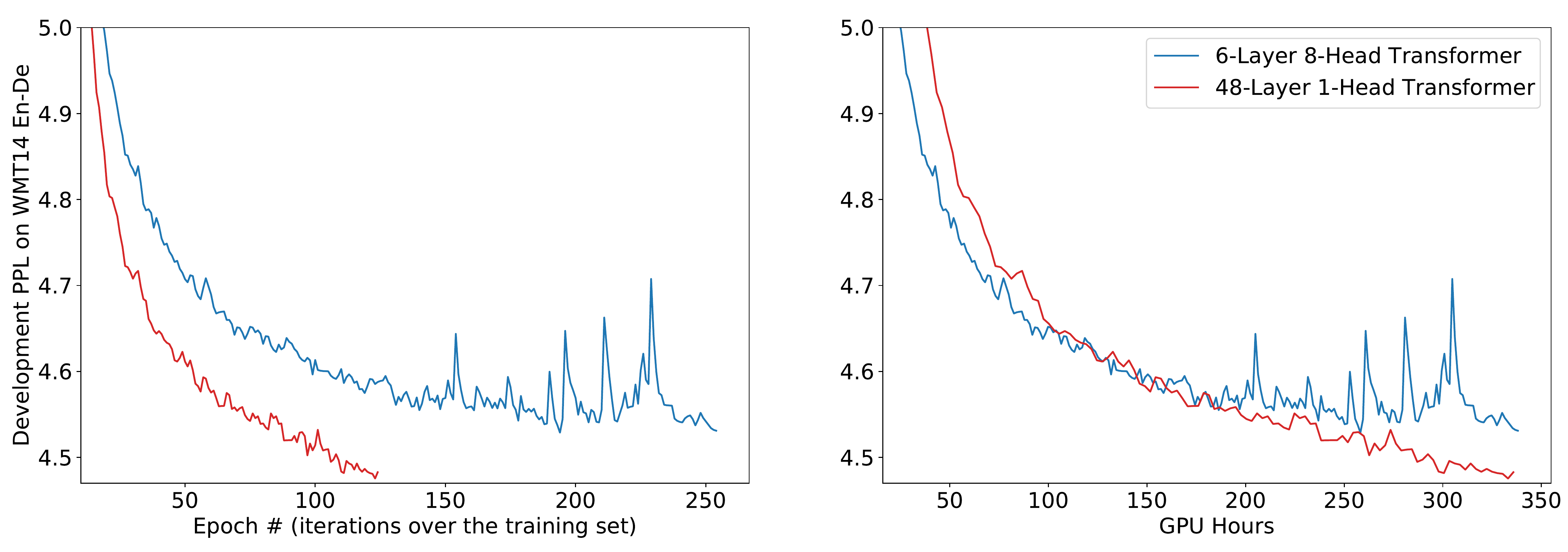}
\caption{Transformer Training Efficiency (GPU Hours are calculated for an idle RTX 3060).}
\label{fig:learning-curve-vs-wall-clock}
\end{figure}

\subsection{Performance Improvements with Different Number of Heads}

Intuitively, the difference between deep single-head Transformer and shallow multi-head Transformer is proportional to the head number (e.g., the difference between 2H-6L and 1H-12L should be smaller than the difference between 4H-6L and 1H-24L). 
Thus, we conduct experiments on Transformers with different head numbers, i.e., $\alpha$H-6L-6L ($\alpha$=1, 2, 4, 6, 8 and feedforward network dimension is $\alpha \cdot 256$) and 1H-$\beta$L-$\beta$L ($\beta$=6,12,24,36,48) for machine translation, $\alpha$H-12L ($\alpha$=1, 3, 6, 12 and feedforward network dimension is $\alpha \cdot 256$) and 1H-$\beta$L ($\beta$=12,36,72,144) for BERT pre-training. 
Note that when $\alpha$ and $\beta$ are set to the smallest value, the architectures of $\alpha$H-6L-6L and 1H-$\beta$L-$\beta$L (or $\alpha$H-12L and 1H-$\beta$L) are the same. 
The performance of deep single-head Transformer and shallow multi-head Transformer are visualized in Figure~\ref{fig:model_size}, and verifies our intuition that since the architecture difference is between shallow multi-head Transformer and deep single-head Transformer is larger for more number of heads, the performance improvement is also larger for more number of heads.

\subsection{GPU Inference Speed Comparison}

The shallow multi-head Transformer and the deep single-head Transformer have roughly the same model size and the computation complexity (as in Table~\ref{tbl:nmt} and \ref{tbl:mnli_squad}). 
Here, we try to verify that they also have similar efficiency in practice. 

Here, we conduct experiments based on the Nvidia FasterTransformer implementation\footnote{We used the version 4.0 as in https://github.com/NVIDIA/FasterTransformer}. 
We calculated the average inference speed on an idle RTX 3060 GPU, and summarized the results in Figure~\ref{fig:inference-speed}. 
It shows that the inference efficiency of the shallow multi-head Transformer and the deep single-head Transformer are roughly the same. 
Specifically, the shallow multi-head Transformer is slightly faster when the batch size and sequence length are smaller, and the deep single-head Transformer is slightly faster when the batch size and the sequence length are larger. 

\subsection{Performance v.s. Wall-Clock Time}

Although the deep single-head Transformer and the shallow multi-head Transformer have roughly the same inference efficiency, we found the training computation of the multi-head Transformer is faster in practice. 
We further conduct experiments to empirically compare the convergence speed of the deep single-head Transformer and the shallow multi-head Transformer. 

As in Figure~\ref{fig:learning-curve-vs-wall-clock}, we can find that the training computation speed of 1H-48L-48L Transformer is about two times slower than the 8H-6L Transformer. 
Meanwhile, the 8H-6L Transformer converges faster with regard to epoch number, or GPU hours. 
This phenomenon verifies our intuition that the network depth of the 6-Layer Transformer has become a bottleneck of the model capacity, which restricts the model performance. 



\section{Conclusion}

Here, we focus on understanding the effectiveness of the multi-head Transformer. 
We first show that, deep single-head Transformer also attends multiple positions and is more effective than the popular shallow multi-head Transformer. 
Then, we suggest the main advantage of the multi-head attention is the training stability, since it has less number of layers than the single-head attention, when attending the same number of positions.
We also show that, with recent advances in deep learning, the training stability is no longer an obstacle and it can lead to consistent performance improvements by turning shallow single-head Transformer to deep multi-head Transformer. 

Our work opens up new possibilities to not only further push the state-of-the-art but understand the effectiveness of Transformer better.
It leads to various interesting future work.
For example, intuitively, both shallow multi-head Transformer and deep single-head Transformer should not be the optimal architecture, and neural architecture search can be employed to find a good balance between the multi-head structure and the single-head structure.

\bibliography{reference}
\bibliographystyle{plainnat}


\appendix

\section*{Appendix}

\subsection*{Implementation Detail}
Besides the layer number and head number, we adopted all hyper-parameters from previous work. 
Specifically, we followed \citep{Liu2020UnderstandingTD}  for machine translation experiments and \citep{Devlin2019BERTPO} for language model pre-training experiments. 
It is worth mentioning that, in \citep{Liu2020UnderstandingTD}, the default initialization method is the Xavier initialization~\citep{Glorot2010UnderstandingTD}, which depends on the size of the weight matrix. 
Here, to control variables, we fix the initialization scale to be the same with original multi-head shallow Transformer. 
Meanwhile, for language model pre-training, since \citep{Devlin2019BERTPO} fixes the initialization scale for all models, we directly adopt the initialization strategy without modification. 

\subsection*{Training Detail}
For machine translation experiments, we followed \citep{Liu2020UnderstandingTD} to conduct data pre-processing, conduct model training on Nvidia GPUs (including Quadro RTX 8000, GeForce RTX 3060, and Quadro RTX A6000). 
As to language model pre-training experiments, we followed \citep{Devlin2019BERTPO} to conduct data pre-processing, conduct model training with Google TPU v3.


\end{document}